# New Approach of Estimating PSNR-B For De-blocked Images

K.Silpa[1], Dr.S.Aruna Mastani [2]

[1] M.Tech (DECS,)Department of ECE, JNTU College of Engineering, Anantapur, Andhra Pradesh, India
Email: k.shilpa410@gmail.com,

[2] Assistant Professor, Department of ECE, JNTU College of Engineering, Anantapur, Andhra Pradesh, India
Email:aruna.mastani@yahoo.com

*Abstract:* **Measurement of image quality is very crucial to many image processing applications. Quality metrics are used to measure the quality of improvement in the images after they are processed and compared with the original images. Compression is one of the applications where it is required to monitor the quality of decompressed / decoded image. JPEG compression is the lossy compression which is most prevalent technique for image codecs. But it suffers from blocking artifacts. Various deblocking filters are used to reduce blocking artifacts. The efficiency of deblocking filters which improves visual signals degraded by blocking artifacts from compression will also be studied. Objective quality metrics like PSNR, SSIM, and PSNR-B for analyzing the quality of deblocked images will be studied. We introduce a new approach of PSNR-B for analyzing quality of deblocked images. Simulation results show that new approach of PSNR-B called "modified PSNR-B" gives even better results compared to existing well known blockiness specific indices.**

*Index Terms*—--- Blocking artifacts, Deblocked images, Quality assessment, Quantization, Quality metrics.

## I. Introduction

Digital images are subject to a wide variety of distortions during acquisition, processing, compression, storage, transmission and reproduction, any of which may result in a degradation of visual quality. Many practical and commercial systems use digital image compression when it is required to transmit or store the image over network bandwidth limited resources. JPEG compression is the most popular image compression standard among all the members of lossy compression standards family. JPEG image coding is based on block based discrete cosine transform. BDCT coding has been successfully used in image and video compression applications due to its energy compacting property and relative ease of implementation.

Blocking effects are common in block-based image and video compression systems. Blocking artifacts are more serious at low bit rates, where network bandwidths are limited. Significant research has been done on blocking artifact reduction [7]–[13]. After segmenting an image in to blocks of size N×N, the blocks are independently DCT transformed, quantized, coded and transmitted. One of the most noticeable degradation of the block transform coding is the "blocking artifact". These artifacts appear as a regular pattern of visible block boundaries. In order to achieve high compression rates using BTC (Block Transform Coding) with visually acceptable results, a procedure known as deblocking is done in order to eliminate blocking artifacts. A deblocking filter can improve image quality in some aspects, but can reduce image quality in other regards.

We perform simulations on deblocked images for analyzing the quality of it. We first perform simulations using the conventional peak signal to noise ratio quality metric, structural similarity index metric. The PSNR does not capture the subjective quality well when blocking artifacts are present. The SSIM metric is slightly more complex the PSNR, but correlates highly with human subjectively. PSNR-B is a quality metric which includes PSNR and a blocking effect factor. While calculating blocking effect factor, the proposed PSNR-B is modified by considering a set of diagonal pixel pairs which are not lying on a block boundary. By going through simulation results, it is shown that new concept of PSNR-B gives better results than the well known blockiness specific index.

Section II reviews lossy compression, deblocking algorithms and change in distortion concept. Section III reviews quality metrics which have been proposed in the literature. In section IV we propose a new approach of PSNR-B quality metric to analyze the quality of deblocked images. Section V presents the simulation results and comparisions. Concluding remarks are presented in section VI.

## II. Quantization and Deblocking Filters

*A) Lossy Compression:*

Quantization is a key element of lossy compression, but information is lost. The amount of compression and the quality can be controlled by the quantization step. As quantization step increases, the quality of the image degrades due to the increase in compression ratio. The trade off exists between compression ratio and deblocked images. The input image is divided into L×L blocks in block transform coding in which each block is transformed independently in to transform coefficients. Therefore an input image block 'b' is transformed into a DCT coefficient block is given by

$$B = TbT^t \quad (2.1)$$

Where T is the transform matrix and $T^t$ is the transpose matrix of T. The transform coefficients are then quantized using a scalar quantizer Q

$$\tilde{B} = Q(B) = Q(TbT^t) \quad (2.2)$$

The quantized coefficients are stored or transmitted to decoder. Therefore the output of the decoder is then given by

$$\tilde{b} = T^t \tilde{B} T = T^t Q(TbT^t)T \quad (2.3)$$

Quantization step is represented by $\Delta$. The SSIM index captures the similarity of reference and test images. As the quantization step size becomes larger, the structural differences between reference and test image will generally increase. Hence, the SSIM index and PSNR are monotonically decreasing functions of the quantization step size $\Delta$.

*B) Deblocking:*

To remove blocking effect, several deblocking techniques have been proposed in the literature as post process mechanisms after JPEG compression. If deblocking is viewed as an estimation problem, the simplest solution is probably just to low pass the blocky JPEG compressed image. The advantage of low pass filtering technique is that no additional information is needed and as a result, the bit rate is not increased. However, it results in blurred images. More sophisticated methods involve iterative methods such as projection on convex sets [3, 4] and constrained least squares [4, 5]. We use deblocking algorithms including low pass filtering and projection on to convex sets. The efficiency of these algorithms and performance of new quality approach can be analyzed by introducing a proposed method in the following sections.

*C) Concept of change in distortion:*

Deblocking operation is performed in order to reduce blocking artifacts. Deblocking operation can be achieved by using various deblocking algorithms, employing deblocking filters. The effects of deblocking filters can be analyzed by introducing a change in distortion concept. The deblocking operation results in the enhancement of image quality in some areas, while degrading in other areas.

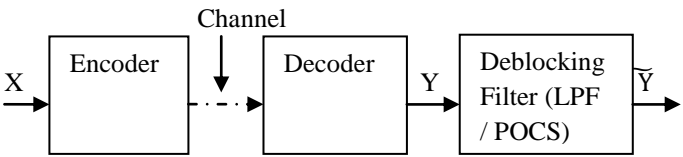

Figure1: Block diagram showing JPEG compression

X – Original Image    Y – Compressed/ Decoded Image    $\tilde{Y}$- Deblocked Image

Let X be the reference image and Y be the test image (decoded image) distorted by quantization errors and $\tilde{Y}$ be the deblocked image as shown in figure1. Let f represent the deblocking operation and is given by $\tilde{Y}=f(Y)$. Let the quality metric between X and Y be M(X,Y). For the given image Y, the main aim of deblocking operation f is to maximize M(X, f(Y)). Let $\alpha_i$ represent the amount of decease in distortion in the decrease in distortion region (DDR) and is given by

$$\alpha_i = d(x_i, y_i) - d(x_i, \tilde{y}_i) \quad (2.4)$$

Where $d(x_i, y_i)$ the distortion between $i^{th}$ pixels of X and Y and is expressed as squared Euclidian distance

$$d(x_i, y_i) = \|x_i - y_i\|^2 \quad (2.5)$$

Where $d(x_i, \tilde{y}_i)$ the distortion between $i^{th}$ pixels of X and Y and is expressed as squared Euclidian distance. Next, we define the distortion decrease region (DDR) to be composed of those pixels where the distortion is decreased by the deblocking operation

$$i \epsilon A, \text{ if } d(x_i, \tilde{y}_i) < d(x_i, y_i) \quad (2.6)$$

The amount of distortion decrease for the $i^{th}$ pixel $\alpha_i$ in the DDRA is

$$\alpha_i = d(x_i, y_i) - d(x_i, \tilde{y}_i) \quad (2.7)$$

We define the mean distortion decrease (MDD)

$$\bar{\alpha} = \frac{1}{N} \sum_{i \in A} (d(x_i, y_i) - d(x_i, \tilde{y}_i)) \quad (2.8)$$

The distortion may also increase at other pixels by application of the deblocking filter. We similarly define the distortion increase region (DIR)B

$$i \epsilon B, \text{ if } d(x_i, y_i) < d(x_i, \tilde{y}_i) \quad (2.9)$$

The amount of distortion increase for the ith pixel $\beta_i$ in the DIRB is

$$\beta_i = d(x_i, \tilde{y}_i) - d(x_i, y_i) \quad (2.10)$$

Where N is the number of pixels in the image. Similarly the mean distortion increase (MDI) is

$$\bar{\beta} = \frac{1}{N} \sum_{i \in B} (d(x_i, \tilde{y}_i) - d(x_i, y_i)) \quad (2.11)$$

The difference between MDD and MDI can be represented as Mean distortion change (MDC) and is given by

$$\bar{\gamma} = \bar{\alpha} - \bar{\beta} \quad (2.12)$$

From this it can be stated that the deblocking operation is likely successful if $\bar{\gamma} > 0$. This is because the mean distortion decrease is larger than the mean distortion increase. Nevertheless, the level of perceptual improvement or loss does not meet these conditions. Based on these conditions, the effect of deblocking filters can be analyzed.

*A) Low pass filter:* A simple L×L low pass deblocking filter can be represented as

$$g(N(x_i)) = \sum_{k=1}^{L^2} h_k \cdot x_{i,k} \quad (2.13)$$

Where $N(x_i)$ represent Neighborhood of pixel $x_i$
'g' represents deblocking operation function
'$h_k$' represents Kernel for the L×L filter
$x_{i,k}$ represents the $k^{th}$ pixel in the L×L neighborhood
of pixel

While low pass filter is used as deblocking filter to reduce blocking artifacts, the distortion will decrease for some pixels defined by (DDR-A) and the distortion will likely increase for some pixels defined by (DIR-B) and it is possible that $\gamma < 0$ could result. The image will be degraded due to blurring as critical high frequency is lost.

*B) POCS:* Deblocking algorithms based upon projection into convex sets (POCS) have demonstrated good performance for reducing blocking artifacts and have proved popular [9]-[13-14]. In POCS Projection operation is done in the DCT domain and low pass filtering operation is done in the spatial domain. Forward DCT and inverse DCT operations are required because the low pass filtering and the projection operations are performed in various domains. Convergence require Multiple iterations and the low pass filtering, DCT, Projection, IDCT operations require one iteration. POCS filtered images converge to an image that does not exhibit blocking artifacts under certain conditions [9], [12], [13]. But computational complexity is more as it requires more iterations.

### III. Existing Quality metrics

To Measure the quality degradation of an available distorted image with reference to the original image, a class of quality assessment metrics called full reference (FR) are considered. Full reference metrics perform distortion measures having full access to the original image. The quality assessment metrics are estimated as follows

*A) PSNR[13][14] :* Peak Signal-to-Noise Ratio (PSNR) and mean Square error are most widely used full reference (FR) QA metrics [2], [13]. As before X is the reference image and Y is the test image. The error signal between X and Y is assumed as 'e'. Then

$$MSE(X,Y) = \frac{1}{N}\sum_{i=1}^{N} e_i^2 = \frac{1}{N}\sum_{i=1}^{N}(x_i - y_i)2 \quad (3.1)$$

$$PSNR(X,Y) = 10 log_{10} \frac{255^2}{MSE(X,Y)} \quad (3.2)$$

Where N represent Number of pixels in an image. However, The PSNR does not correlate well with perceived visual Quality [14], [15]-[18].

*B) SSIM [9]:* The Structural similarity (SSIM) metric aims to measure quality by capturing the similarity of images [2]. Three aspects of similarity: Luminance, contrast and structure is determined and their product is measured. Luminance comparison function $l(X,Y)$ for reference image X and test image Y is defined as below

$$l(X,Y) = \frac{2\mu_X \mu_Y + C1}{\mu_x^2 + \mu_y^2 + C1} \quad (3.3)$$

Where $\mu_x$ and $\mu_y$ are the mean values of X and Y respectively and C1 is the stabilization constant.
Similarly the contrast comparison function $c(X, Y)$ is defined as

$$c(X,Y) = \frac{2\sigma_x \sigma_y + C2}{\sigma_x^2 + \sigma_y^2 + C2} \quad (3.4)$$

Where the standard deviation of X and Y are represented as $\sigma_x$ and $\sigma_y$ and C2 is the stabilization constant.
The structure comparison function $s(X, Y)$ is defined as

$$s(X,Y) = \frac{\sigma_{xy} + C3}{\sigma_x \sigma_y + C3} \quad (3.5)$$

Where $\sigma_{xy}$ represents correlation between X and Y and $C_3$ is a constant that provides stability.
By combining the three comparison functions, The SSIM index is obtained as below

$$SSIM(X,Y) = [l(X,Y)]^\alpha \cdot [c(X,Y)]^\beta \cdot [s(X,Y)]^\gamma \quad (3.6)$$

and the parameters are set as $\alpha = \beta = \gamma = 1$ and C3=C2/2

From the above parameters the SSIM index can be defined as

$$SSIM(X,Y) = \frac{(2\mu_X \mu_Y + C1)(2\sigma_{xy} + C2)}{(\mu_x^2 + \mu_y^2 + C1)(\sigma_x^2 + \sigma_y^2 + C2)} \quad (3.7)$$

Symmetric Gaussian weighting functions are used to estimate local SSIM statics. The mean SSIM index pools the spatial SSIM values to evaluate overall image quality [2].

$$SSIM(X,Y) = \frac{1}{M}\sum_{j=1}^{M} SSIM(x_j - y_j) \quad (3.8)$$

Where $x_j$ and $y_j$ are image patches covered by the j[th] window and the number of local windows over the image are represented by M.

*C) PSNR-B [ 14] :* PSNR-B is a quality metric which is specifically used for measuring the quality of images which consists of blocking artifacts. As that of other metrics it includes Peak Signal-to-Noise Ratio (PSNR) and in addition a blocking effect factor (BEF) which measures blockiness of images. Generally the blocking artifacts is a problem during compression where the original image is required to be divided into sub images called blocks. So this metric is effectively used in assessing the quality of decompression/deblocked images.

In this quality metric, the BEF is calculated by considering horizontal and vertical neighboring pixel pairs which are not lying across block boundaries. But this may not include the artifacts that occurs in the diagonal directions at the boundaries. In order to consider this ,we included a BEF with diagonal neighboring pixel pairs along with BEF of the horizontal and vertical neighboring pixel pairs. However in the course of our experimentation with many decompressed images it is found that BEF using only diagonal approach (diagonal neighboring pixels) is more effective than the existing horizontal approach PSNR-B (horizontal neighboring pixels). It is also observed that the proposed diagonal approach called '**Modified PSNR-B**' gives the same result as that of combined BEF approach ( horizontal and diagonal) . The detailed concept of proposed method will be discussed in next section.

Consider an image that contains integer number of blocks such that the horizontal and vertical dimensions of the image are divisible by block dimension and the blocking artifacts occur along the horizontal and vertical dimensions[14].

| $Y_1$ | $Y_9$ | $Y_{17}$ | $Y_{25}$ | $Y_{33}$ | $Y_{41}$ | $Y_{49}$ | $Y_{57}$ |
|---|---|---|---|---|---|---|---|
| $Y_2$ | $Y_{10}$ | $Y_{18}$ | $Y_{26}$ | $Y_{34}$ | $Y_{42}$ | $Y_{50}$ | $Y_{58}$ |
| $Y_3$ | $Y_{11}$ | $Y_{19}$ | $Y_{27}$ | $Y_{35}$ | $Y_{43}$ | $Y_{51}$ | $Y_{59}$ |
| $Y_4$ | $Y_{12}$ | $Y_{20}$ | $Y_{28}$ | $Y_{36}$ | $Y_{44}$ | $Y_{52}$ | $Y_{60}$ |
| $Y_5$ | $Y_{13}$ | $Y_{21}$ | $Y_{29}$ | $Y_{37}$ | $Y_{45}$ | $Y_{53}$ | $Y_{61}$ |
| $Y_6$ | $Y_{14}$ | $Y_{22}$ | $Y_{30}$ | $Y_{38}$ | $Y_{46}$ | $Y_{54}$ | $Y_{62}$ |
| $Y_7$ | $Y_{15}$ | $Y_{23}$ | $Y_{31}$ | $Y_{39}$ | $Y_{47}$ | $Y_{55}$ | $Y_{63}$ |
| $Y_8$ | $Y_{16}$ | $Y_{24}$ | $Y_{32}$ | $Y_{40}$ | $Y_{48}$ | $Y_{56}$ | $Y_{64}$ |

Figure2: Example for illustration of pixel blocks
The blocking effect factor specifically measures the amount of blocking artifacts just using the test image. It can be defined as

$$BEF(Y) = \eta[D_B(Y) - D_B^C(Y)] \qquad (3.9)$$

Where $D_B(Y)$ = mean boundary pixel squared difference of test Image and $D_B^C(Y)$ = mean nonboundary pixel squared difference of test Image by considering a set of horizontal and vertical neighboring pixel pairs which are not lying on a block boundary.

Where

$$\eta = \begin{cases} \dfrac{\log_2^B}{\log_2^{(\min(N_H,N_V))}} & , \text{ if } D_B(Y) > D_B^C(Y) \\ 0 & , \text{ otherwise} \end{cases} \quad (3.10)$$

The mean square error including blocking effects for reference image X and test image Y is defined as follows,

$$MSE - B(x, y) = MSE(x, y) + BEF_{Tot}(y) \qquad (3.11)$$

Where $BEF_{Tot}(Y) = \sum_{k=1}^{K} BEF_k(y)$ (3.12)

Finally the existing PSNR-B is given as,

$$PSNR - B(x, y) = 10 \log_{10} \frac{255^2}{MSE - B(x,y)} \qquad (3.13)$$

## IV. Proposed method: Modified PSNR-B

*Modified PSNR-B:* A new quality metric called modified PSNR -B, which is same as proposed one but here we are considering set of diagonal neighboring pixel pairs which are not lying across block boundaries along with the horizontal and vertical neighboring pixel pairs. Consider an image that contains integer number of blocks such that the horizontal and vertical dimensions of the image are divisible by block dimension and the blocking artifacts occur along the horizontal, vertical and diagonal dimensions.

| $Y_1$ | $Y_9$ | $Y_{17}$ | $Y_{25}$ | $Y_{33}$ | $Y_{41}$ | $Y_{49}$ | $Y_{57}$ |
|---|---|---|---|---|---|---|---|
| $Y_2$ | $Y_{10}$ | $Y_{18}$ | $Y_{26}$ | $Y_{34}$ | $Y_{42}$ | $Y_{50}$ | $Y_{58}$ |
| $Y_3$ | $Y_{11}$ | $Y_{19}$ | $Y_{27}$ | $Y_{35}$ | $Y_{43}$ | $Y_{51}$ | $Y_{59}$ |
| $Y_4$ | $Y_{12}$ | $Y_{20}$ | $Y_{28}$ | $Y_{36}$ | $Y_{44}$ | $Y_{52}$ | $Y_{60}$ |
| $Y_5$ | $Y_{13}$ | $Y_{21}$ | $Y_{29}$ | $Y_{37}$ | $Y_{45}$ | $Y_{53}$ | $Y_{61}$ |
| $Y_6$ | $Y_{14}$ | $Y_{22}$ | $Y_{30}$ | $Y_{38}$ | $Y_{46}$ | $Y_{54}$ | $Y_{62}$ |
| $Y_7$ | $Y_{15}$ | $Y_{23}$ | $Y_{31}$ | $Y_{39}$ | $Y_{47}$ | $Y_{55}$ | $Y_{63}$ |
| $Y_8$ | $Y_{16}$ | $Y_{24}$ | $Y_{32}$ | $Y_{40}$ | $Y_{48}$ | $Y_{56}$ | $Y_{64}$ |

Figure3: Example for illustration of pixel blocks

Let $N_H$ and $N_v$ be the horizontal and vertical dimensions of the $N_H X N_v$ image I. Let $\mathcal{H}$ be the set of horizontal neighboring pixel pairs in I. Let $\mathcal{H}_B \subset \mathcal{H}$ be the set of horizontal neighboring pixel pairs that lie across a block boundary. Let $R_B^C$ be the set of right sided diagonal neighboring pixel pairs, not lying across a block boundary, i.e. $R_B^C = \mathcal{H} - \mathcal{H}_B$, . Similarly, let $v$ be the set of vertical neighboring pixel pairs, and $v_B$ be the set of vertical neighboring pixel pairs lying across block boundaries. Let $L_B^C$ be the set of left sided diagonal neighboring pixel pairs not lying across block boundaries i.e. $L_B^C = v - v_B$.

$$N_{H_B} = N_V \left(\frac{N_H}{B}\right) - 1 \qquad (4.1)$$

$$N_{R_B^C} = N_V(N_H - 1) - N_{H_B} \qquad (4.2)$$

$$N_{V_B} = N_H \left(\frac{N_V}{B}\right) - 1 \qquad (4.3)$$

$$N_{L_B^C} = N_H(N_V - 1) - N_{V_B} \qquad (4.4)$$

Where $N_{H_B}, N_{H_B^C}, N_{V_B}, N_{V_B^C}$ be the number of pixel pairs in $\mathcal{H}_B, \mathcal{H}_B^C, v_B$ and $v_B^C$ respectively and B is the block size.

Fig. 2 shows a simple example for illustration of pixel blocks with $N_H = 8, N_V = 8$, and B=4. The thick lines represent the block boundaries. In this example $N_{H_B} = 8$, $N_{H_B^C} = 48$, $N_{V_B} = 8$, and $N_{V_B^C} = 48$. The sets of pixel pairs in this example are

$\mathcal{H}_B = \{(y_{25}, y_{33}), (y_{26}, y_{34}), \ldots (y_{32}, y_{40})\}$ (a)
$\mathcal{H}_B^C = \{y_1, y_9), (y_9, y_{17}), (y_{17}, y_{25}), \ldots (y_{56}, y_{64})\}$ (b)
$v_B = \{(y_4, y_5), (y_{12}, y_{13}), \ldots (y_{60}, y_{61})\}$ (c)
$v_B^C = (y_1, y_2), (y_2, y_3), (y_3, y_4), (y_5, y_6), \ldots (y_{63}, y_{64})\}$ (d)
(4.5)

Fig. 3 shows a simple example for illustration of pixel blocks with $N_H = 8, N_V = 8$, and B=4. The thick lines represent the block boundaries. In this example $N_{H_B} = 8$, $N_{L_B^C} = 48$, $N_{V_B} = 8$, and $N_{R_B^C} = 48$. The sets of pixel pairs in this example are

$\mathcal{H}_B = \{(y_{25}, y_{33}), (y_{26}, y_{34}), \ldots (y_{32}, y_{40})\}$ (a)
$R_B^C = \{y_1, y_{10}), (y_9, y_{18}), (y_{17}, y_{26}), \ldots (y_{55}, y_{64})\}$ (b)
$v_B = \{(y_4, y_5), (y_{12}, y_{13}), \ldots (y_{60}, y_{61})\}$ (c)
$L_B^C = (y_9, y_2), (y_{17}, y_{10}), (y_{25}, y_{18}), (y_{41}, y_{34}), \ldots (y_{63}, y_{56})\}$ (d)
(4.6)

Then we define the mean boundary pixel squared difference $(D_B)$ and the mean nonboundary pixel squared difference $(D_{B_C})$ for image y to be

$$D_B(Y) = \frac{\sum_{(y_i,y_j) \in \mathcal{H}_B}(y_i - y_j)^2 + \sum_{(y_i,y_j) \in v_B}(y_i - y_j)^2}{N_{H_B} + N_{V_B}} \qquad (4.7)$$

$$D_B^C(Y) = \frac{\sum_{(y_i,y_j) \in R_B^C}(y_i - y_j)^2 + \sum_{(y_i,y_j) \in L_B^C}(y_i - y_j)^2}{N_{R_B^C} + N_{L_B^C}} \qquad (a)$$

The above equation is applicable if only diagonal neighboring pixel pairs are considered.

$$D_B^C(Y) =$$

$$\frac{\sum_{(y_i,y_j) \in \mathcal{H}_B^C}(y_i-y_j)^2 + \sum_{(y_i,y_j) \in v_B^C}(y_i-y_j)^2 + \sum_{(y_i,y_j) \in R_B^C}(y_i-y_j)^2 + \sum_{(y_i,y_j) \in L_B^C}(y_i-y_j)^2}{2*(N_{H_B^C} + N_{V_B^C})}$$

(b) (4.8)

If we consider all combination of pixel pairs include horizontal, vertical and diagonal neighboring pixel pairs, equation 4.8(b) is applicable. Blocking artifacts will become more visible as the quantization step size increases; mean boundary pixel squared difference will increase relative to mean non boundary pixel square difference. The blocking effect factor is given by

$$BEF(Y) = \eta[D_B(Y) - D_B^C(Y)] \qquad (4.9)$$

Where

$$\eta = \begin{cases} \frac{log_2^B}{log_2^{(\min(N_H, N_V))}} & , \text{ if } D_B(Y) > D_B^C(Y) \quad (4.10) \\ 0 & , \text{ otherwise} \end{cases}$$

A decoded image may contain multiple block sizes like 16×16 macro block sizes and 4×4 transform blocks, both contributing to blocking effects. Then the blocking effect factor for $k^{th}$ block is given by

$$BEF_k(Y) = \eta_k [D_{B_k}(Y) - D_{B_k}^C(Y)] \quad (4.11)$$

For overall block sizes BEF is given by

$$BEF_{Tot}(Y) = \sum_{k=1}^{K} BEF_k(y) \quad (4.12)$$

The mean square error including blocking effects for reference image X and test image Y is defined as follows,

$$MSE\_B(x,y) = MSE(x,y) + BEF_{Tot}(y) \quad (4.13)$$

Finally the proposed PSNR-B is given as,

$$PSNR\_B(x,y) = 10 log_{10} \frac{255^2}{MSE\_B(x,y)} \quad (4.14)$$

The MSE measures the distortion between the reference image and the test image, while the BEF specifically measures the amount of blocking artifacts just using the test image. These no-reference quality indices claim to be efficient for measuring the amount of blockiness, but may not be efficient for measuring image quality relative to full-reference quality assessment. On the other hand, the MSE is not specific to blocking effects, which can substantially affect subjective quality. We argue that the combination of MSE and BEF is an effective measurement for quality assessment considering both the distortions from the original image and the blocking effects in the test image. The associated quality index PSNR-B is obtained from the MSE-B by a logarithmic function, as is the PSNR from the MSE. The PSNR-B is attractive since it is specific for assessing image quality, specifically the severity of blocking artifacts. The modified PSNR-B produces even better results compared to the PSNR-B, PSNR and other well known blockiness specific index. It is computationally efficient.

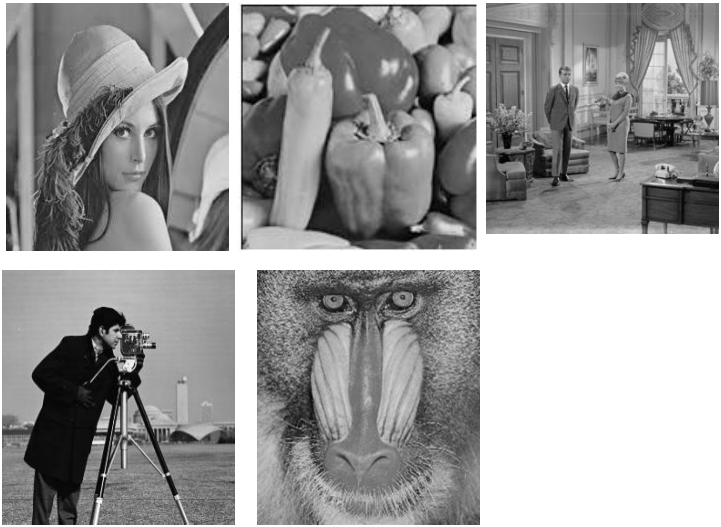

Figure 4: database images (a) Lena image (b) Peppers image (c) Leopard image (d) Cameraman image (e) Mandril image

**V. Simulation Results on Deblocked images:**

In this paper image quality assessment is done by objective measurement in which evaluations are automatic and mathematical defined algorithms. Generally, Quality metrics are used to measure the quality of improvement in the images after they are processed and compared with the original images. For experimentation, ata base images JPEG Compression is one of the applications where it is required to monitor the quality of decompressed / decoded image, The impact blocking artifacts is . a serious problem in JPEG compression/decompression model. Thus Various deblocking filters ( are used to reduce blocking artifacts and resulting deblocked images are compared w.r.t various quality metrics ( ). The comparison of quality metrics is also made by varying the quantization step size. The images of USC-SIPI[ ] database is used for experimentation. Some of the Sample images of this database over which the quality metrics are compared are as shown in the fig.4. comparison of quality metrics for the above images is illustrated graphically from fig.5 to fig.9. From these graphs, It is observed that the proposed quality metric "modified PSNR_B" gives best performance compared to the existing metrics(). A detailed analysis of the graphical result (fig.6) for one of the images (pepper ig.4) is discussed here.

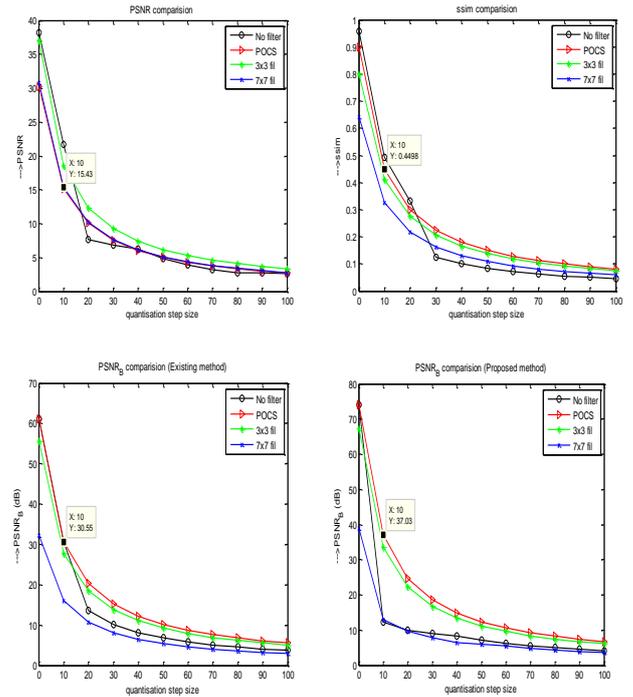

Figure 5: Comparison of quality metrics for Lena image
(a) PSNR (b) SSIM (c) PSNR-B (d) modified PSNR-B

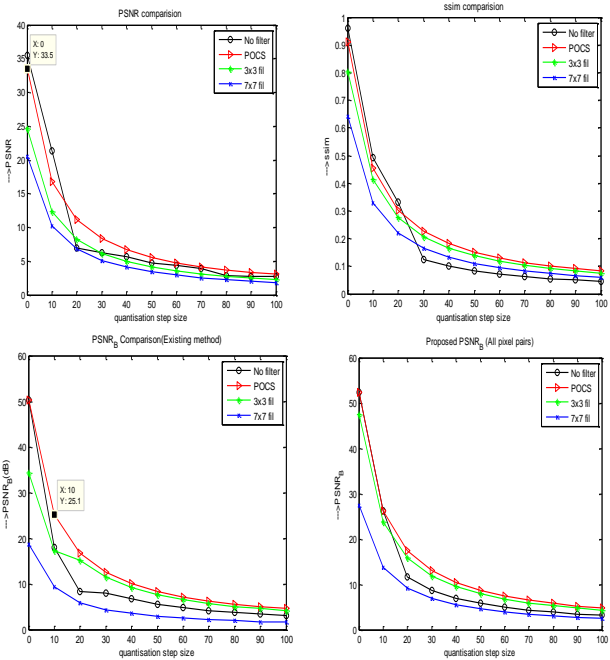

Figure 6 : Comparison of quality metrics for Peppers image (a) PSNR (b) SSIM (c) PSNR-B (d) modified PSNR-B

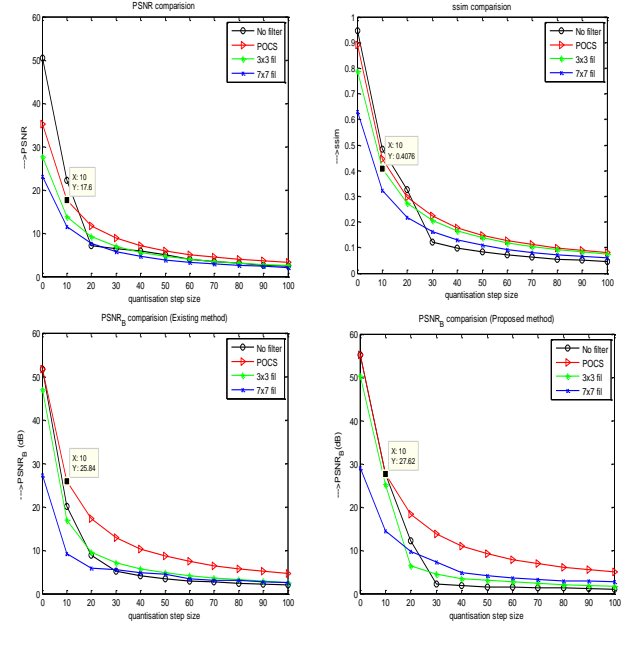

Figure 8: Comparison of quality metrics for cameraman image (a) PSNR (b) SSIM (c) PSNR-B(d)modified PSNR-B

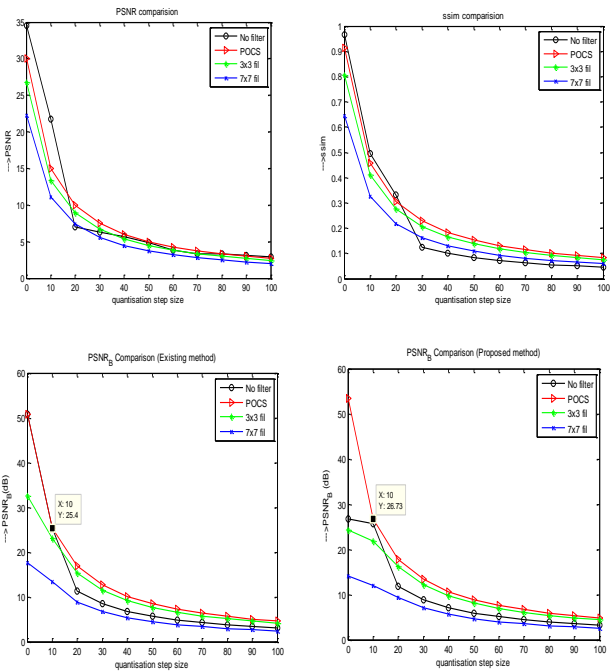

Figure7: Comparison of quality metrics for Living Room image (a) PSNR (b) SSIM (c) PSNR-B (d) modified PSNR-B

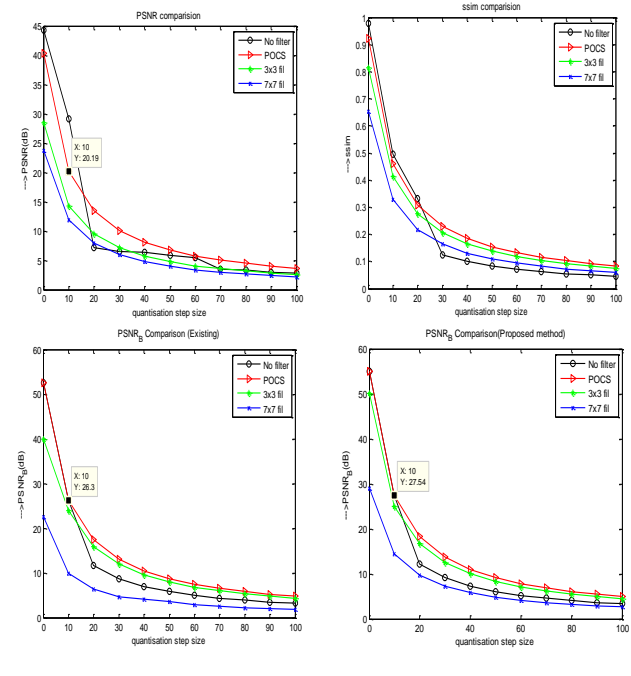

Figure 9: Comparison of quality metrics for Mandril image (a) PSNR (b) SSIM (c) PSNR-B(d)modified PSNR-B

**Comparison of quality metrics**: Simulations are performed on these image and quality metrics are estimated. Quantization step sizes of 10, 20, 30, 40, 50, and 100 are used in the simulations to analyze the effects of quantization step size

*A) PSNR Analysis:*

Fig. 6 – (a) shows that when the quantization step size was large ($\Delta \geq 20$), the no filter, 3×3 filter, and POCS methods resulted in higher PSNR than the 7×7 filter case on the image. All the deblocking methods produced lower PSNR when the quantization step size was small ($\Delta \leq 20$).

*B) SSIM Analysis:*

Fig. 6-(b) shows that when the quantization step was large ($\Delta \geq 20$), on the image, all the filtered methods resulted in larger SSIM values. The 3×3 and 7×7 low pass filters resulted in lower SSIM values than the no filter case when the quantization step size was small ($\Delta \leq 30$).

*C) PSNR-B Analysis:*

Fig. 6 – (c) shows that when the quantization step size was large ($\Delta \geq 10$), the no filter, 7×7 filter, and POCS methods resulted in higher PSNR than the 3×3 filter case on the image. All the deblocking methods except POCS produced lower PSNR when the quantization step size was small ($\Delta \leq 20$).

*D) Modified PSNR-B Analysis:*

Fig.6 – (d) shows that when the quantization step size was large ($\Delta \geq 10$), the no filter, 7×7 filter, and POCS methods resulted in higher PSNR than the 3×3 filter case on the image. Comparing to PSNR-B, a new concept of modified PSNR-B produced better results for all quantization steps.

## VI. CONCLUSIONS

Image quality assessment plays an important role in various image processing applications. Experimental results indicate that MSE and PSNR are very simple, easy to implement and have low computational complexities. But these methods do not show good results. MSE and PSNR are acceptable for image similarity measure only when the images differ by simply increasing distortion of a certain type. But they fail to capture image quality when they are used to measure across distortion types. SSIM is widely used method for measurement of image quality. It works accurately can measure better across distortion types as compared to MSE and PSNR, but fails in case of highly blurred image. Natural images and standard images were tested by these quality metrics. Those sample images are shown in above figure. We have found that PSNR-B is the better quality metric for JPEG compression which shows better performance than the other well known quality metrics. Similarly a new approach of PSNR-B produced even better results compared to the proposed PSNR-B.

For future work, we look forward to new problems in this direction of inquiry. Firstly, quality studies of PSNR-B and perceptually proven index SSIM in conjunction are of considerable value, not only for studying deblocking operations, but also for other image improvement applications, such as restoration, denoising, enhancement, and so on. Post processing methods using POCS have shown good performance for blocking artifact reduction. The iterative operations in POCS require infeasible amount of computations for practical real time applications. DPOCS (DCT domain POCS) is a post process technique and it is an efficient non-iterative post processing method. The proposed method can even be extended to color images and videos.

## REFERENCES

[1] Y. Yang, N. P. Galatsanos, and A. K. Katsaggelos, "Projection-based spatially adaptive reconstruction of block-transform compressed images," *IEEE Trans. Image Process.*, vol. 4, no. 7, pp. 896–908, Jul. 1995.

[2] Y. Yang, N. P. Galatsanos, and A. K. Katsaggelos, "Regularized reconstruction to reduce blocking artifacts of block discrete cosine transform compressed images," *IEEE Trans. Circuits Syst. Video Technol.*, vol. 3, no. 6, pp. 421–432, Dec. 1993.

[3] H. Paek, R.-C. Kim, and S. U. Lee, "On the POCS-based post processing technique to reduce the blocking artifacts in transform coded images," *IEEE Trans.Circuits Syst. Video Technol.*, vol. 8, no. 3, pp. 358–367, 1998.

[4] S. H. Park and D. S. Kim, "Theory of projection onto narrow quantization constraint set and its applications," *IEEE Trans. Image Process.*, vol. 8, no. 10, pp. 1361–1373, Oct. 1999.

[5] A. Zakhor, "Iterative procedure for reduction of blocking effects in transform image coding," *IEEE Trans. Circuits Syst. Video Technol.*, vol. 2, no. 1, pp. 91–95, Mar. 1992.

[6] S. Liu and A. C. Bovik, "Efficient DCT-domain blind measurement and reduction of blocking artifacts," *IEEE Trans. Circuits Syst. Video Technol.*, vol. 12, no. 12, pp. 1139–1149, Dec. 2002.

[7] Z.Wang and A. C. Bovik, "Blind measurement of blocking artifacts in images," in *Proc. IEEE Int. Conf. Image Process.*, Vancouver, Canada, Oct. 2000, pp. 981–984.

[8] Y.Yang, N.P.Galatsanos, and A.K.Katsaggelos, "*Regularized reconstruction to reduce blocking artifacts of block discrete cosine transform compressed images,*" IEEE Trans. Circuits Syst. Video Technol., vol.3, no.6, pp.421-432, Dec.1993.

[9] Z.Wang, A.C.Bovik, and E.P.Simoncelli, "*Multi-scale structural similarity for image quality assessment*," in Proc. IEEE Asilomar Conf.Signal Syst. Comput.,No v.2003.

[10] Y.Jeong, I.Kim, and H.Kang," *Practical projection based postprocessing of block coded images with fast convergence rate,*" IEEE Trans. Circuits Syst. Video Technol., vol.10,no.4 , pp.617-623,Jun.2000.

[11] P. List, A. Joch, J. Laimena, J. Bjøntegaard, and M. Karczewicz, "Adaptive deblocking filter," *IEEE Trans. Circuits Syst. Video Technol.*, vol. 13, no. 7, pp. 614–619, Jul. 2003.

[12] A. M. Eskicioglu and P. S. Fisher, "Image quality measures and their performance," *IEEE Trans. Communications*, vol. 43, pp. 2959–2965, Dec. 1995.

[13] G. Zhai, W. Zhang, X. Yang, W. Lin, and Y. Xu, "No-reference noticeable blockiness estimation in images," *Signal Process. Image Commun.*, vol. 23, pp. 417–432, 2008.

[14] Changhoon Yim,Member IEEE and Alan Conrad Bovik, fellow IEEE "Quality Assessment of deblocked images " *IEEE Transactions on image processin.,* vol. 20, No.1, 2011.